%% file: main.tex
\documentclass[10pt,twocolumn,letterpaper]{article}

\usepackage[pagenumbers]{wacv} 
\usepackage{amsmath}
\usepackage{booktabs}
\usepackage{multirow}


\definecolor{wacvblue}{rgb}{0.21,0.49,0.74}
\usepackage[pagebackref,breaklinks,colorlinks,allcolors=wacvblue]{hyperref}

\def\wacvPaperID{2383} 
\def\confName{WACV}
\def\confYear{2026}

\title{SCAdapter: Content-Style Disentanglement for Diffusion Style Transfer}

\author{
Luan Thanh Trinh \quad Kenji Doi \quad Atsuki Osanai\\
LY Corporation\\
{\tt\small \{ttrinh, kedoi, atsuki.osanai\}@lycorp.co.jp}
}

\begin{document}

\maketitle
\input{sec/0_abstract}    
\begin{figure*}
  \centering
  \includegraphics[width=\textwidth]{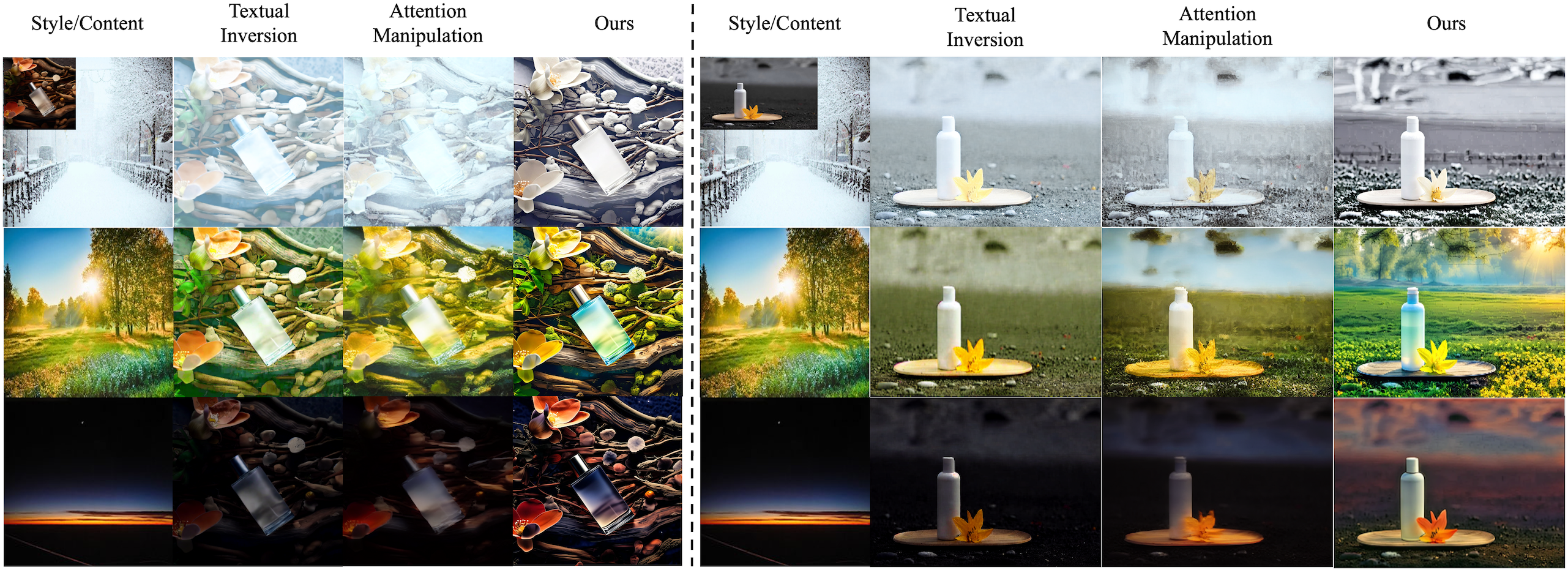}
  \caption{Our method surpasses diffusion-based style transfer approaches like textual inversion (InST) and attention manipulation (StyleID) by modifying both composition and fine details (shadows, lighting, snowflakes). Competitors often produce painting-like results or miss detailed stylistic elements in photorealistic transformations.}
  \label{fig:Qualitative_compare_photo}
  \vspace{-10pt}
\end{figure*}
\input{sec/1_intro}
\input{sec/2_relatedwork}
\input{sec/3_method}

\input{sec/4_exp}
\input{sec/5_conclu}
{

    \small
    \bibliographystyle{ieeenat_fullname}
    \bibliography{my}
}

\end{document}


\maketitle

\section{Additional applications - Text-driven stylized synthesis.}
\label{sec:additional_applications}
Text-driven stylized synthesis is a highly sought-after feature that has attracted significant research interest. In this task, given a text prompt and a style reference image, users aim to generate images that embody similar styles. The proposed method, with its ability to extract pure style features from the style reference image, is promising for delivering good results in text-driven stylized synthesis.

\begin{figure*}[t]
    \centering
    \includegraphics[width=0.7\linewidth]{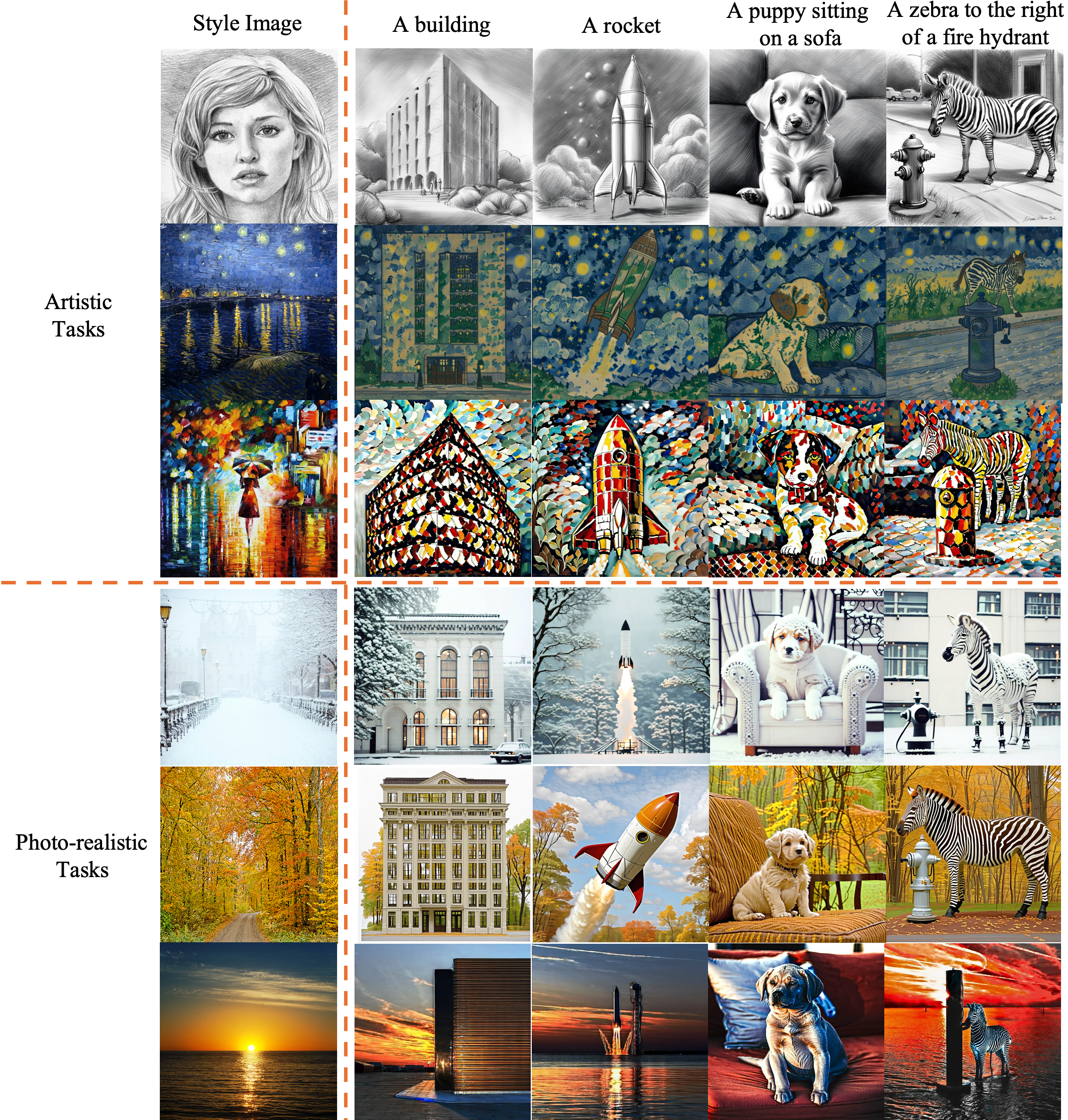}
    \caption{SCAdapter delivers high-quality text-driven stylized synthesis for both artistic and photo-realistic tasks.}
    \label{fig:t2i}
    
\end{figure*}

To adapt our approach for this task, we removed the content extractor branch from SCAdapter. Fig.~\ref{fig:t2i} showcases example results where our method demonstrates strong alignment with both reference styles and input prompts. For an objective evaluation, we conducted a quantitative comparison with several prominent models in text-driven stylized synthesis, as shown in Tab.~\ref{tab:t2i_comparision}. We assessed text alignment capability (TA) by measuring cosine similarity within the CLIP text-image embedding space between the textual prompts and their corresponding synthesized images. To evaluate style similarity, we utilized SFS from Eq.~13.

The quantitative results in Tab.~\ref{tab:t2i_comparision} indicate that, although not specifically designed for this task, the proposed method performs competitively when compared to other methods. Notably, it achieves highest TA scores, indicating a strong alignment between input prompts and output content. This success is due to the extraction of pure style features from the style reference image, effectively minimizing unnecessary influences from its content features.

\begin{table}[ht]
    \vspace{-5pt}
    \centering
    \caption{Quantitative comparison with other text-driven stylized synthesis baselines.}
    \resizebox{\columnwidth}{!}{%
    \begin{tabular}{lcccccc}
    \toprule
    \textbf{Method} & \textbf{InST} & \textbf{CAST} & \textbf{StyTR2} & \textbf{T2I-Adapter} & \textbf{DEADiff} & \textbf{Ours} \\
    \midrule
    TA & 0.237 & 0.282 & 0.282 & 0.224 & \underline{0.284} & \textbf{0.285} \\
    SFS & 0.238 & 0.244 & 0.234 & \textbf{0.271} & 0.259 & \underline{0.266} \\
    \bottomrule
    \end{tabular}%
    }
    \vspace{-5pt}
    \label{tab:t2i_comparision}
\end{table}

\section{Qualitative comparison with ablation of KVS Injection and Style Consistency Loss}
\begin{figure*}[t]
    \centering
    \includegraphics[width=0.5\linewidth]{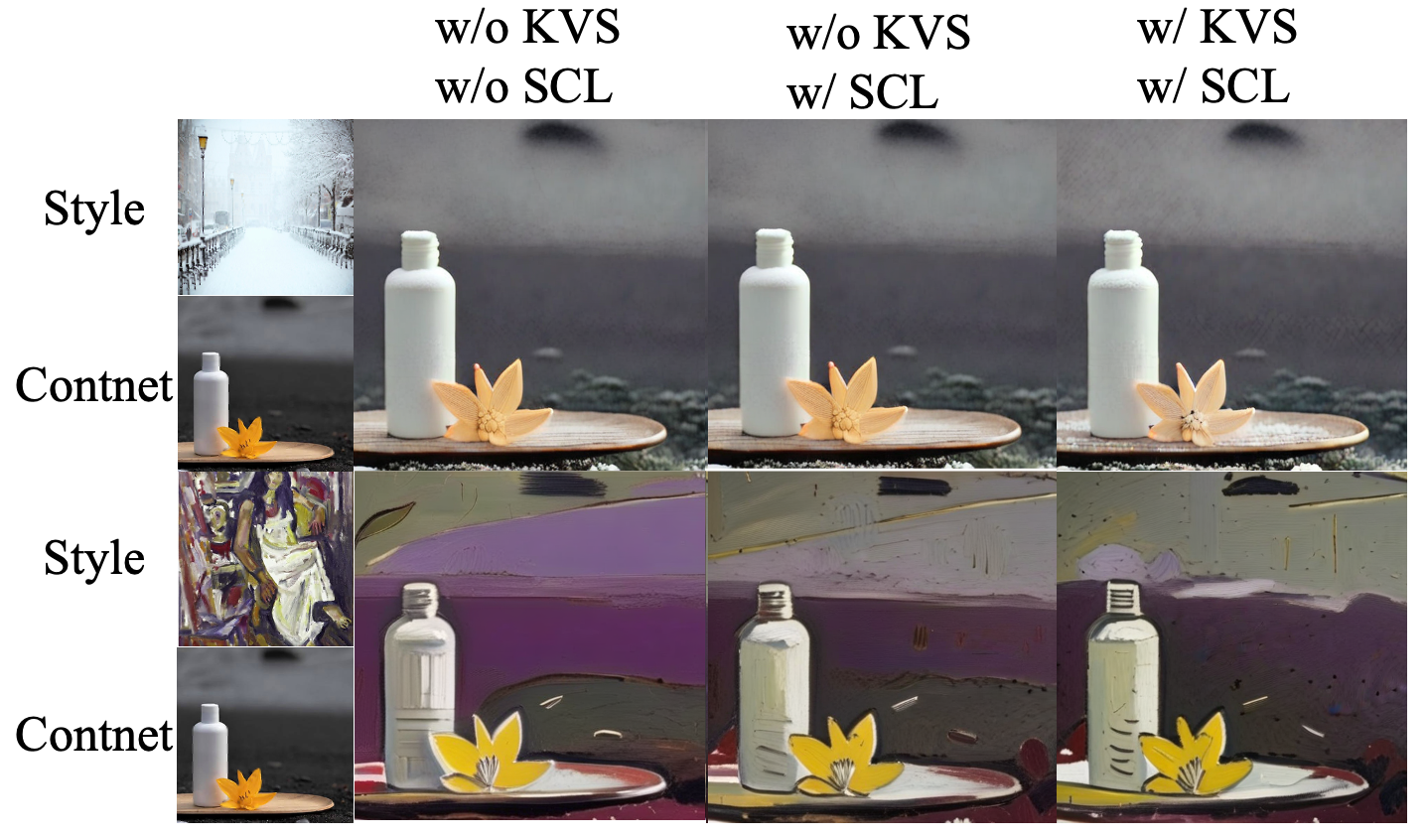}
    \caption{Qualitative comparison while ablating the KVS Injection and Style Consistency Loss.}
    \label{fig:quali_ablation}

\end{figure*}

To highlight the effects of KVS Injection (denoted as KVS in Fig.~\ref{fig:quali_ablation}) and the style consistency loss (denoted as SCL in Fig.~\ref{fig:quali_ablation}), we present qualitative examples of style transfer results obtained by ablating these components. Consistent with the quantitative findings reported in the main paper, both KVS Injection and the style consistency loss contribute to improving the final performance of our proposed model, with KVS Injection playing a particularly significant role. For the photo-realistic transfer task (row 1), snowflake effects are added more abundantly and in a more harmonious manner when KVS Injection is applied. For the artistic transfer task (row 2), the painting style characteristics are expressed more vividly. Meanwhile, although the style consistency loss provides measurable improvements, its impact is less apparent from a qualitative perspective. These observations indicate that the proposed model remains effective when trained on standard image datasets, without the need to rely on a content–style–stylized triplet dataset.

\section{More results of artistic transfer}
\begin{figure*}[t]
    \centering
    \includegraphics[width=\linewidth]{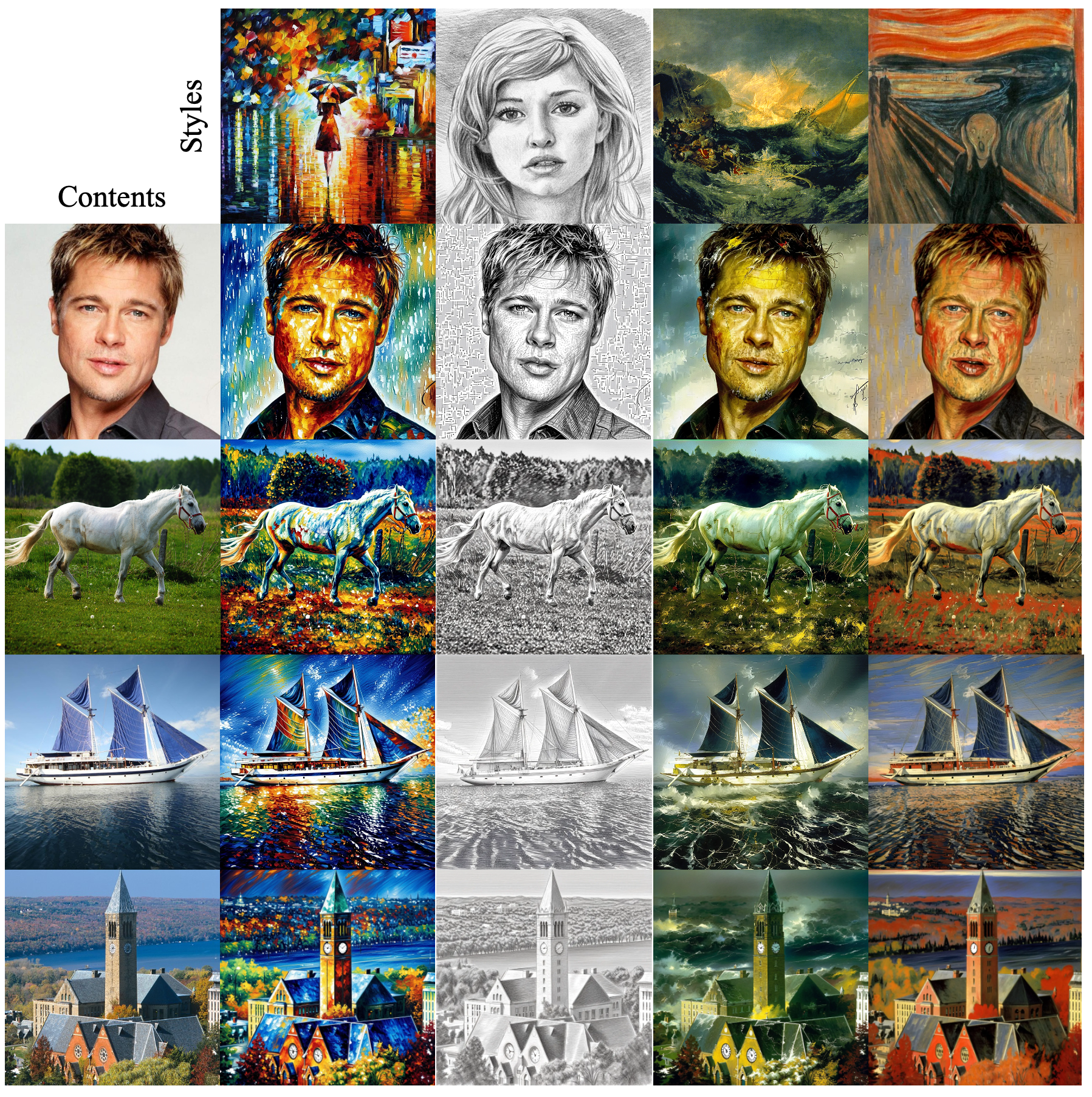}
    \caption{More results of artistic transfer. Zoom in for viewing details.}
    \label{fig:more_art}

\end{figure*}

Fig.~\ref{fig:more_art} presents several additional results of our proposed model applied to artistic style transfer. As shown, the model produces visually compelling outputs that effectively capture the artistic characteristics of the reference styles while preserving the structural content of the original images.

%% file: sec/0_abstract.tex
\begin{abstract}
Diffusion models have emerged as the leading approach for style transfer, yet they struggle with photo-realistic transfers, often producing painting-like results or missing detailed stylistic elements. Current methods inadequately address unwanted influence from original content styles and style reference content features. We introduce SCAdapter, a novel technique leveraging CLIP image space to effectively separate and integrate content and style features. Our key innovation systematically extracts pure content from content images and style elements from style references, ensuring authentic transfers. This approach is enhanced through three components: Controllable Style Adaptive Instance Normalization (CSAdaIN) for precise multi-style blending, KVS Injection for targeted style integration, and a style transfer consistency objective maintaining process coherence. Comprehensive experiments demonstrate SCAdapter significantly outperforms state-of-the-art methods in both conventional and diffusion-based baselines. By eliminating DDIM inversion and inference-stage optimization, our method achieves at least $2\times$ faster inference than other diffusion-based approaches, making it both more effective and efficient for practical applications.
\end{abstract}

%% file: sec/1_intro.tex
\section{Introduction}
\label{sec:intro}

Recent advancements in Diffusion Models (DMs) have significantly transformed generative applications, including text-to-image synthesis \cite{1,2,30} and image/video editing \cite{3,4,5}. These models have proven particularly valuable for style transfer tasks \cite{31,32,33,34}, where style characteristics from a reference image are applied to a content image while preserving its fundamental structure and semantic information.

Among diffusion-based methods, attention manipulation \cite{45, 6, 52} and textual inversion \cite{7, 16} stand out as the most promising for style transfer, though each faces distinct challenges. Attention manipulation methods, such as the recently proposed baseline StyleID \cite{6}, operate on the principle that \textit{query} features from attention layers represent content, while \textit{key} and \textit{value} features encapsulate style elements. By merging \textit{query} features of the content image with \textit{key} and \textit{value} features of the style image, these methods effectively produce stylized images, achieving state-of-the-art (SOTA) results in artistic transfer tasks. However, they encounter significant difficulties with photo-realistic transfer tasks, as photo-realistic styles often lack the distinctive characteristics inherent in artworks. This limitation is particularly evident when altering the style of product images (e.g., normal-to-snowy, day-to-night), as shown in Fig.~\ref{fig:Qualitative_compare_photo}. Textual inversion methods, particularly InST \cite{7} and InstantStyle \cite{16}, leverage prompt embedding spaces to encode target styles, integrating seamlessly with prompts to generate styled images. Though less effective than attention manipulation methods for artistic transfer, visual comparisons in Fig.~\ref{fig:Qualitative_compare_photo} reveal its surprising proficiency in photorealistic style transfer.

A common limitation in diffusion-based methods is their propensity to apply style transfer without fully removing existing style features from content images. They also often overlook the influence of content features from style references, potentially altering the final image composition. Various solutions have been proposed to tackle the content-style disentanglement challenge. DiffuseIT \cite{46} utilizes specialized loss functions for style and content with pre-trained diffusion models via inference-stage optimization. While this approach avoids the need for additional training, the process remains time-consuming for each style-content pair, and our results in Sec.~\ref{subsec:quantitative_compare}  demonstrate its limited performance. StyleDiffusion \cite{12} assumes images with removed styles belong to the same domain, using a surrogate domain and CLIP-based loss to train diffusion models. However, this reliance on surrogate domains and the need for fine-tuning each target style significantly limits its applicability.

To address these fundamental limitations, we propose the Style Content Adapter (SCAdapter) framework—a novel approach that leverages the CLIP \cite{14} image space to systematically extract pure content features from content images and isolate style features from style reference images. Unlike prior methods that depend on surrogate domains or inference-time optimization, our approach offers a more direct and effective mechanism for disentangling content and style. These cleanly separated elements are then mapped into the prompt embedding space via specialized tokenizers and integrated with prompt embeddings, thereby guiding diffusion models to generate stylized images with unprecedented fidelity and control.

\begin{table}[ht]
\centering
\caption{Comparison with textual inversion and attention manipulation approaches.}
\resizebox{\columnwidth}{!}{%
\begin{tabular}{lcc}
\toprule
\textbf{Method} & \textbf{Style Space} & \textbf{Style Injection} \\
\midrule
Textual Inversion & CLIP & To prompt embedding \\
Attention Manipulation & Noise latent & \textit{key, value} (style latent) vs \textit{query} (content latent) \\
Ours & CLIP & \textit{key, value} (style, prompt) vs \textit{query} (content latent) \\
\bottomrule
\end{tabular}%
}
\label{table:compare methodology}
\end{table}

Building on insights from StyleID \cite{6}, which demonstrates that the \textit{key} and \textit{value} components in attention layers encode style characteristics, we introduce KVS Injection—a refined attention-based style injection technique that more effectively incorporates style features. By using KVS Injection, our model can be viewed as a hybrid method that unifies the strengths of the two dominant paradigms: textual inversion and attention manipulation—while additionally employing explicit content-style disentanglement. Tab.~\ref{table:compare methodology} presents a high-level comparison of our approach with these existing methods.

For many practical applications, especially in photorealistic style transfer where partial preservation of the original style is often desirable, we develop Controllable Style Adaptive Instance Normalization (CSAdaIN). This sophisticated weighting mechanism enables precise control over the contribution of original and new styles to the final output. Our experimental results in Sec.~\ref{subsec:abalation} further demonstrate that CSAdaIN can effectively mix multiple target styles rather than relying on a single style reference image, a significant advancement over traditional style transfer methods \cite{6,7,21,22,23,24,25}. To enhance CSAdaIN's capability to manage multiple styles, improve its cohesion with KVS Injection, and boost overall style transfer and content preservation effectiveness, we integrate a style transfer consistency objective within standard diffusion model training. 

Our main contributions are summarized as follows:
\begin{itemize}
    \item We introduce a novel style transfer method leveraging the CLIP image space to distinctly separate and recombine content and style features. By extracting pure content from content images and style from style references, our method avoids cross-contamination and ensure cleaner, more authentic transfers.
    \item We enhance the proposed method by introducing three key components: KVS Injection, style transfer consistency objective, and notably, CSAdaIN—a method to blend and combine multiple styles harmoniously.
    \item Extensive experiments on the style transfer dataset demonstrate that the proposed method significantly outperforms previous methods and achieves state-of-the-art performance. By eliminating the need for DDIM inversion and inference-stage optimization, our approach provides at least $2\times$ the inference speed compared to previous diffusion-based methods, making it both more effective and efficient for practical applications.
\end{itemize}

%% file: sec/2_relatedwork.tex
\section{Related Works}
\label{sec:related work}
\subsection{Diffusion Model-based Neural Style Transfer}
Neural style transfer \cite{35,36,37,38,39} merges the style of one image with the content of another. Within diffusion models, several innovative methods have emerged. InST \cite{7} uses textual inversion to transform styles into textual embeddings, while DiffStyle \cite{32} introduces a training-free approach that exploits h-space and modifies skip connections. Other techniques \cite{31, 47} employ text input to condition either style or content generation. StyleID \cite{6} achieves style transfer by substituting the \textit{key} and \textit{value} features of content with those from style images in self-attention layers.

However, these methods commonly apply style transfer without entirely removing existing style features from content images and overlook content features from style references, potentially altering final composition. To overcome these challenges, we propose using the CLIP image space to separately extract and recombine content and style features, ensuring cleaner transfers by eliminating cross-contamination.

\subsection{Content-Style Disentanglement}
\label{subsec:cs-disentanglement}
Early content-style (C-S) disentanglement relied on labeled data for supervised factorization \cite{8}, though unsupervised methods also exist \cite{9,10}. These often depend on GANs \cite{11}, restricting them to predefined domains. StyleDiffusion \cite{12} uses diffusion models assuming that style-removed images converge into a single domain, pre-training on surrogate domains like photographs. DiffuseIT \cite{46} focuses on inference-stage optimization with specialized loss functions, avoiding additional training but adding computational overhead.

StyleAdapter \cite{27} and DEADiff \cite{28} recently harness the CLIP image space to extract style features from reference images, primarily for text-to-image generation. However, extracting pure content features from content images for style transfer remains unaddressed.

Our method leverages CLIP image space for content-style separation, building on IP-Adapter \cite{13}'s insight that CLIP encoders capture both style and content features in compact vectors. We isolate content features by averaging vectors of identical content in varying styles, then extract style features by subtracting content from overall CLIP embeddings. This enables precise style transfer without surrogate domains or inference-stage optimization, marking a significant departure from previous disentanglement methods.

%% file: sec/3_method.tex
\begin{figure}[t]
    \centering
    \includegraphics[width=\linewidth]{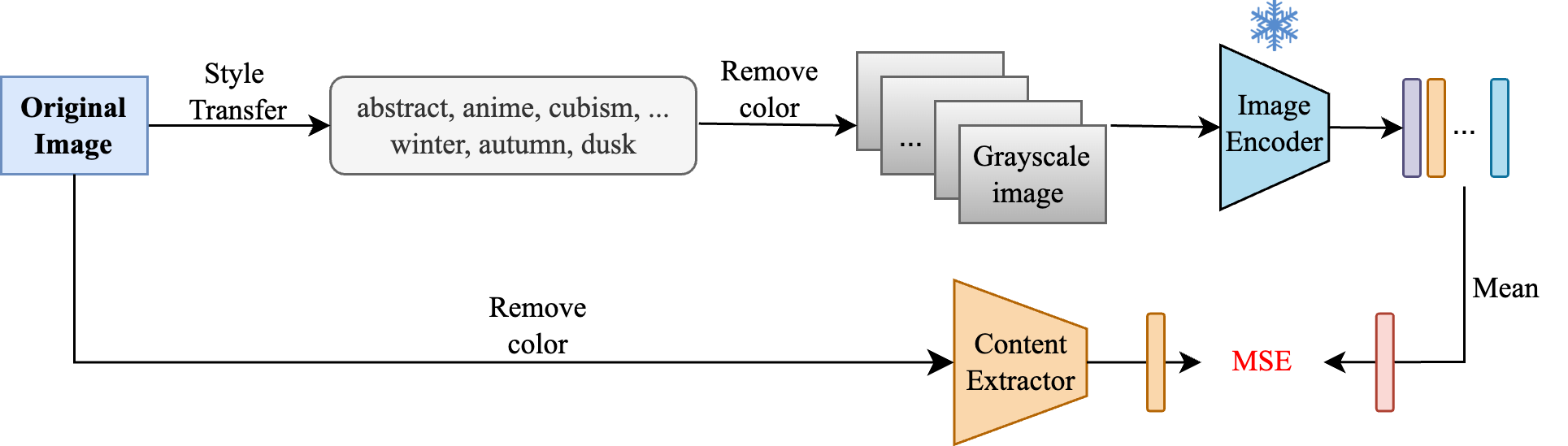}
    \caption{Content Extractor training pipeline}
    \label{fig: Content Extractor}
    \vspace{-15pt}
\end{figure}
\section{Method}
\subsection{Content Extractor and Style Extractor}
\label{subsec:content extractor}
As mentioned in Sec.~\ref{subsec:cs-disentanglement}, style features can be effectively removed by averaging vectors in the CLIP image space that share identical content but exhibit various styles. We propose a pipeline to train such a Content Extractor based on this idea, as illustrated in Fig.~\ref{fig: Content Extractor}.

First, we randomly select 5,000 photo realistic images and 5,000 painting images from the PD12M dataset \cite{15}. Using LoRA and previous style transfer methods, we generate stylized images with the original content across 15 distinct styles: \textit{abstract, anime, cubism, impressionism, modern art, paint, realism, romanticism, colored sketch, sketch, sunshine, winter, autumn, and dusk}.

We employ the CLIP ViT-H/14 \cite{41} model, trained on the LAION-2B English dataset \cite{29}, as the reference and initial model for the content extractor. We fine-tune only CLIP's last visual projector. Given that color is widely acknowledged as a crucial element in defining an image's style \cite{43}, we remove color from the images prior to feature extraction to enhance the effectiveness of content feature extraction. After fine-tuning the content extractor $\boldsymbol{C}$, we can obtain a style extractor $\boldsymbol{S}$ to extract style features from an input image $I$ by utilizing the CLIP Image Encoder as:
\begin{equation}
    \boldsymbol{S}(I) = CLIP(I) - \boldsymbol{C}(I)
    \label{eq: style extractor}
\end{equation}
\subsection{SCAdapter Framework}
\label{subsec:SCAdapter framework}
\begin{figure}[t]
    \centering
    \includegraphics[width=\linewidth]{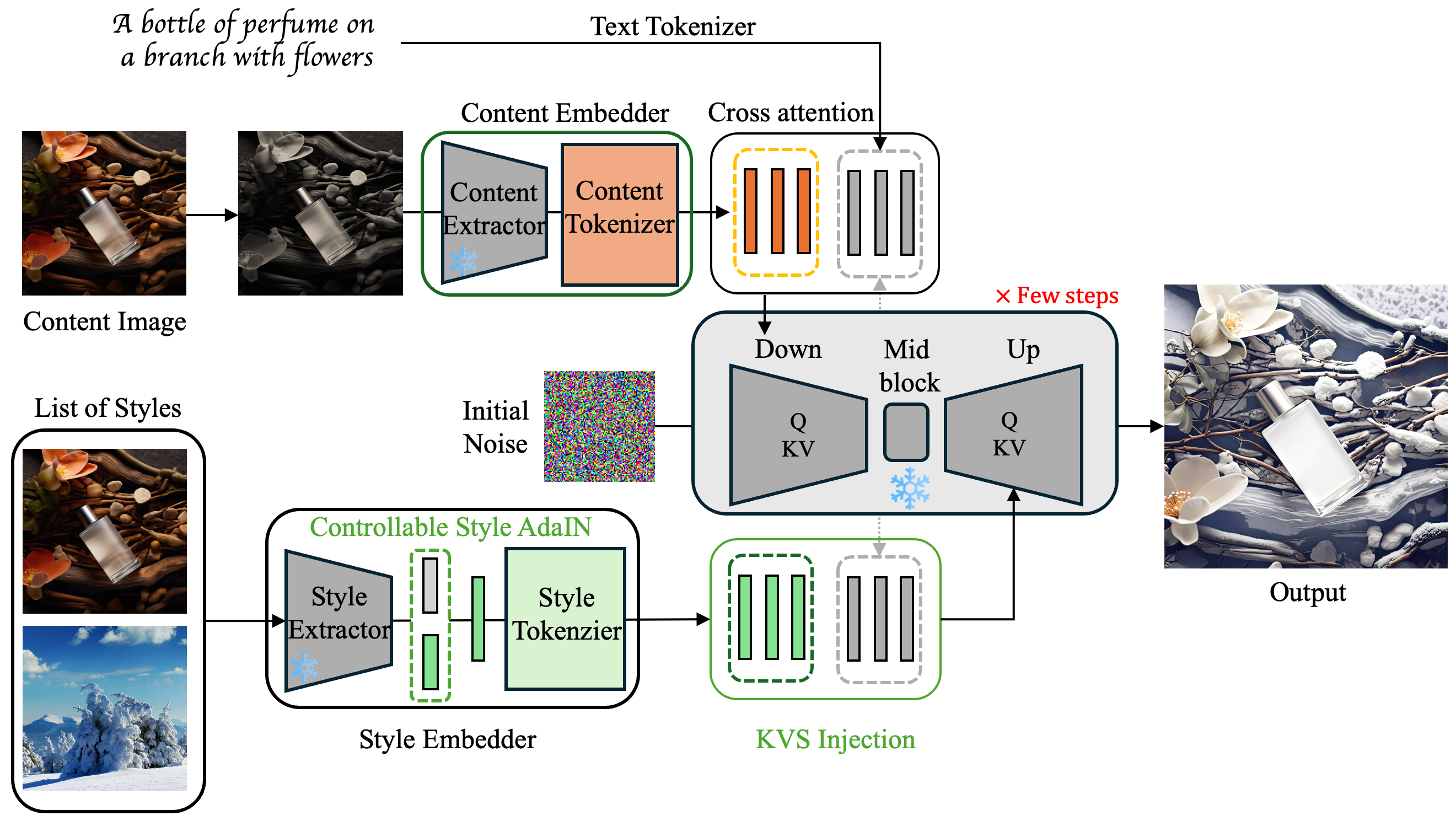}
    \caption{Overview of the proposed SCAdapter framework. }
    \label{fig:SCAdapter framework}
    \vspace{-10pt}
\end{figure}
After obtaining the content and style extractors, we introduce the SCAdapter framework for transferring the style of any image, as illustrated in Fig.~\ref{fig:SCAdapter framework}. Pure content features of the content image, extracted by the content extractor, are transformed into the prompt embedding space through the content tokenizer, which consists of two MLP layers. These features are then combined with the prompt embedding using cross-attention to guide the denoising UNet on the content of the output image. The prompt embedding space often struggles to control the image layout, leading to discrepancies between the content and output images. To address this issue, we utilize Canny or Tile ControlNet \cite{42} during inference. 

\noindent\textbf{Style Combination with CSAdaIN.}\quad In style transfer, particularly in photo-realistic applications, retaining a portion of the content image's original style is crucial. For instance, when changing the style of a product image from summer to winter, or from daylight to sunset, preserving significant aspects of the original style is necessary to avoid altering the product's appearance. Conversely, many artistic transfer scenarios require combining multiple styles rather than relying on a single style reference image, as is common in traditional style transfer methods \cite{6,7,21,22,23,24,25}. To address these challenges, we propose Controllable Style Adaptive Instance Normalization (CSAdaIN). 

Given two style features $S_1$ and $S_2$, we use a weighting parameter $\omega \in [0,1]$ for controllable style interpolation:
\begin{equation}
S_{\text{combined}} = \sigma_{\omega}(S_1, S_2)\left(\frac{S_1 - \mu(S_1)}{\sigma(S_1)}\right) + \mu_{\omega}(S_1, S_2)
\label{eq:contronabble_style_transfer}
\end{equation}
\noindent where the interpolated statistics $\mu_{\omega}$ and $\sigma_{\omega}$ are defined as:
\begin{align}
\mu_{\omega}(S_1, S_2) &= \omega \cdot \mu(S_1) + (1 - \omega) \cdot \mu(S_2)\\
\sigma_{\omega}(S_1, S_t) &= \omega \cdot \sigma(S_1) + (1 - \omega) \cdot \sigma(S_2)
\label{eq:CSAdaIN}
\end{align}
Experimental results in Sec.~\ref{subsec:abalation} demonstrate that this method, despite its simplicity and lack of training requirements, offers high effectiveness and convenience.

\begin{figure}[t]
    \centering
    \includegraphics[width=0.9\linewidth]{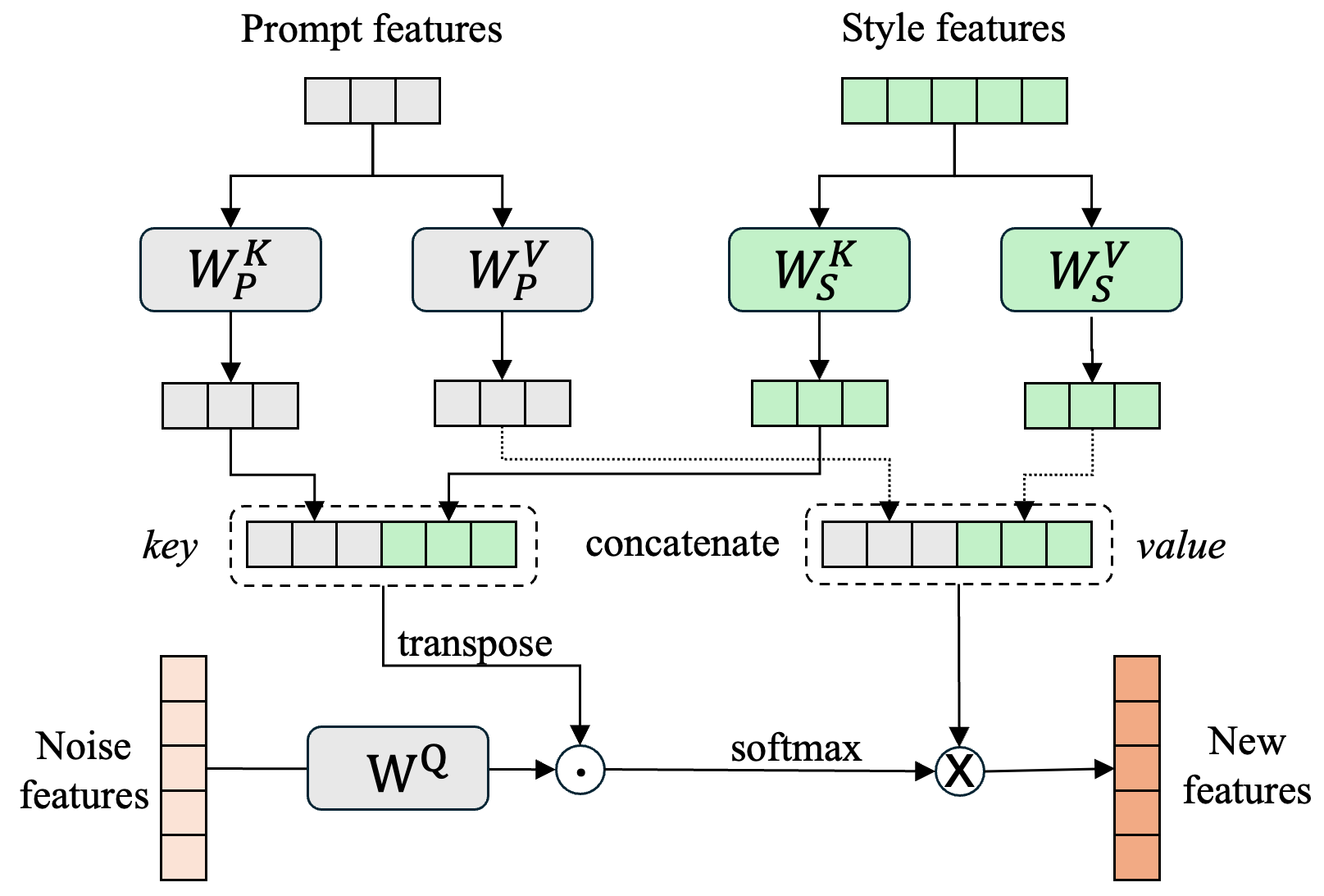}
    \caption{The illustration of KVS Injection.}
    \label{fig: kvstyle}
    \vspace{-10pt}
\end{figure}

\noindent\textbf{Style Injection with KVS Injection.}\quad The next important consideration is how to inject style features into the denoising UNet alongside prompt embeddings to achieve effective style transfer. Our initial experiments suggest that using cross-attention alone is effective primarily for content features. While cross-attention can convey general style characteristics such as color and overall drawing attributes, it struggles to capture small, local style features. To address this issue, we adopt StyleID \cite{6}'s insight that the \textit{key} and \textit{value} features of attention layers capture style features, while \textit{query} features primarily represent content features. Given style features $c_S$ extracted by style extractor and prompt embeddings $c_p$, we compute and concatenate their \textit{key} $K$ and \textit{value} $V$ features. Then, the final features are calculated using these $K$, $V$, and the U-Net query features $Z$. Formally, the formulation of this combination can be expressed as follows:
\begin{align}
    Q &= ZW^Q \\
    K &= \text{Concat}(c_pW^K_P,c_SW^K_S) \\
    V &= \text{Concat}(c_pW^V_P,c_SW^V_S) \\
    Z_{\text{new}} &= \text{Softmax}\left( \frac{QK^T}{\sqrt{d}}\right)V
\end{align}
We refer to this method as Key-Value Style Injection (KVS Injection). A visual illustration of the method is provided in Fig.~\ref{fig: kvstyle}.

To ensure efficiency and reduce computational cost, we strategically inject features into the denoising UNet. Based on findings from \cite{16}, where upsample block attention influences style and downsample block attention affects content, we inject content features—combined with prompt embeddings—into downsample blocks, and style features into upsample blocks. This also enhances content-style disentanglement.

\begin{figure}[t]
    \centering
    \includegraphics[width=\linewidth]{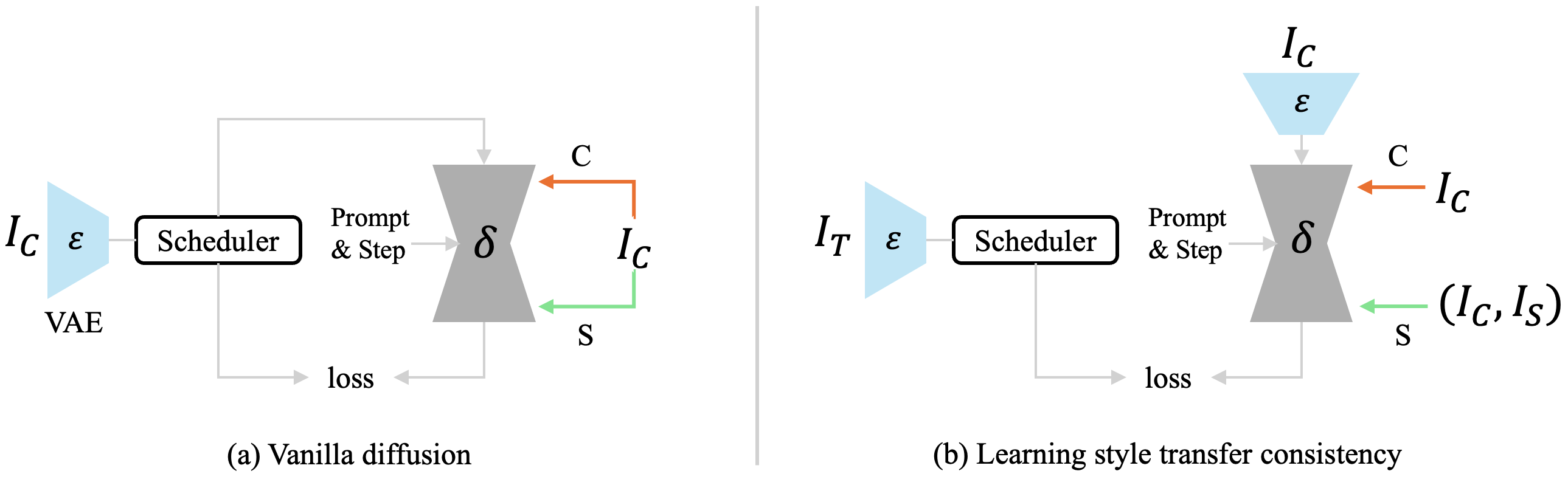}
    \caption{Learning objective.}
    \label{fig: learning objective}
    \vspace{-10pt}
\end{figure}

\subsection{Learning objective}
\label{subsec: model training and inference}
\noindent\textbf{Vanilla objective.} \quad As shown in Fig.~\ref{fig:SCAdapter framework}, we are left with only four components that need fine-tuning: the content tokenizer, style tokenizer, cross-attention block, and KVS Injection block. With content extractor and style extractor trained to extract pure content and style features, the training of the remaining components does not have any special requirements regarding the dataset or algorithm. It can be accomplished using a standard image dataset and the vanilla objectives of diffusion models, as shown in Fig.~\ref{fig: learning objective}-a. 

Given an input content image $\boldsymbol{I_C}$, latent diffusion algorithms first encode $\boldsymbol{I_C}$ as a latent image $\boldsymbol{\epsilon}(\boldsymbol{I_C})$ and then progressively add noise to produce a noisy latent $\boldsymbol{\epsilon}(\boldsymbol{I_C})_t$, where $t$ represents the number of noise additions.
Considering the set of conditions including timestep $t$ and prompt $p$, 
 our model uses content extractor \textit{\textbf{C}} and style extractor \textit{\textbf{S}} to extract content features and style features from the content images to predict the added noise:
\begin{equation}
\mathcal{L}_{\text{vanilla}} = \left| \epsilon - \delta(\boldsymbol{\epsilon}(\boldsymbol{I_C})_t, t, p, \boldsymbol{C}(\boldsymbol{I_C}), \boldsymbol{S}(\boldsymbol{I_C}) \right|_2^2
\end{equation}
Here, $\omega$ in Eq. 3 and 4 is set to 0 to utilize style and content features of the content image and to reconstruct the content image itself. 


\noindent\textbf{Style transfer consistency objective.}\quad The experimental results in Sec.~\ref{subsec:abalation} demonstrate that using only the vanilla objective is effective for typical style transfer tasks. However, to i) enhance CSAdaIN's ability to blend and control multiple styles, ii) improve cohesion between CSAdaIN and KVS Injection, and iii) boost the overall effectiveness of style transfer and content preservation, we introduce a style transfer consistency objective alongside the vanilla diffusion model objective for training.

As illustrated in Fig.~\ref{fig: learning objective}-b, we utilize style-transferred images $\boldsymbol{I_T}$ created by LoRA and other style transfer methods discussed in Sec.~\ref{subsec:content extractor}. The style features from the content image $\textbf{S}(I_C)$ and style reference image $\textbf{S}(I_S)$ are combined using CSAdaIN to create style guidance for the denoising UNet. The model uses this guidance, along with content features, prompt embeddings, and noisy latent $\boldsymbol{\epsilon}(\boldsymbol{I_C})_t$, to reconstruct $\boldsymbol{I_T}$. The formal formulation of this objective can be expressed as follows:
\begin{equation}
\mathcal{L}_{\text{consistency}} = \left| \epsilon - \delta(\boldsymbol{\epsilon}(\boldsymbol{I_C})_t, t, p, \boldsymbol{C}(\boldsymbol{I_C}), \boldsymbol{S}(\boldsymbol{I_C,I_S}) \right|_2^2
\end{equation}
\noindent\textbf{Joint learning objective.}\quad The final learning objective can be written as below with $\lambda=0.1$ as default weights. During training, we set it up to randomly use $\mathcal{L}_{\text{consistency}}$ with a probability of 30\%.
\begin{equation}
    \mathcal{L} = \mathcal{L}_{\text{vanilla}} + \lambda\mathcal{L}_{\text{consistency}}
\end{equation}
\vspace{-20pt}

\begin{table*}
\centering
\renewcommand{\arraystretch}{1.2}
\setlength{\tabcolsep}{3pt}
\resizebox{0.95\textwidth}{!}{%
\begin{tabular}{c c c c c c c c c c c c c}
\toprule
\multirow{2}{*}{\textbf{Task}} & \multirow{2}{*}{\textbf{Metric}} & \multicolumn{6}{c}{\textbf{Diffusion model baselines}} & \multicolumn{5}{c}{\textbf{Conventional baselines}} \\
\cmidrule(lr){3-8} \cmidrule(lr){9-13}
 &  & \begin{tabular}[c]{@{}c@{}}\textbf{Ours}\\(SDXL)\end{tabular} & \begin{tabular}[c]{@{}c@{}}\textbf{Ours}\\(SD1.5)\end{tabular} & \begin{tabular}[c]{@{}c@{}}\textbf{StyleID}\end{tabular} & \begin{tabular}[c]{@{}c@{}}\textbf{AttenST}\end{tabular} & \begin{tabular}[c]{@{}c@{}}\textbf{DiffuseIT}\end{tabular} & \begin{tabular}[c]{@{}c@{}}\textbf{InST}\end{tabular} & \begin{tabular}[c]{@{}c@{}}\textbf{InstantStyle}\end{tabular} & \begin{tabular}[c]{@{}c@{}}\textbf{DiffStyle}\end{tabular} & \begin{tabular}[c]{@{}c@{}}\textbf{AesPA-Net}$^2$\end{tabular} & \begin{tabular}[c]{@{}c@{}}\textbf{CAST}\end{tabular} & \begin{tabular}[c]{@{}c@{}}\textbf{StyTR2}\end{tabular} \\
\midrule
\multirow{5}{*}{\begin{tabular}[c]{@{}c@{}}Photo-realistic\\transfer\end{tabular}} & $\text{FID}_{style}$  $\downarrow$& \textbf{21.151} & \underline{22.417} & 24.349 & 25.176 & 35.314  & 23.048 & 23.051 & 33.119 & 33.166 & 32.951 & 30.998  \\
 & $\text{LPIPS}_{content}$ $\downarrow$ & \textbf{0.3015} & \underline{0.3057} & 0.3774 & 0.3781 & 0.5581 & 0.4185 & 0.4202 & 0.5226 & 0.4172 & 0.6121 & 0.5514 \\
 & Aesthetic Score $\uparrow$ & \textbf{5.5467} & \underline{5.2198} & 4.3193 & 4.3017 & 4.4371 & 4.8512 & 4.7173 & 4.7912 & 4.9173 & 4.7714 & 4.6975  \\
 \cmidrule(lr){2-13}
 & CS $\uparrow$ & \textbf{0.8991} & \underline{0.8813} & 0.8176 & 0.8097 & 0.7983 & 0.8667 & 0.8619 & 0.8017 & 0.8079 & 0.8103 & 0.8139  \\
 & SS $\uparrow$ & \textbf{0.2913} & \underline{0.2873} & 0.2292 & 0.2287& 0.2057 & 0.2331 & 0.2329 & 0.1979 & 0.1991 & 0.2010 & 0.2193 \\
\midrule
\multirow{5}{*}{\begin{tabular}[c]{@{}c@{}}Artistic\\transfer\end{tabular}} & $\text{FID}_{style}$  $\downarrow$& \textbf{18.105} & 18.201 & \underline{18.131} & 19.091 & 23.065 & 21.571 & 21.413 & 20.903 & 19.760 & 20.395 & 18.890  \\
 & $\text{LPIPS}_{content}$ $\downarrow$& \textbf{0.4939} & \underline{0.4951} & 0.5055 & 0.5102 & 0.6921 & 0.8002 & 0.8117 & 0.8931 & 0.5135 & 0.6212 & 0.5445  \\
 & ArtFID $\downarrow$& \textbf{28.541} & \underline{28.707} & 28.801 & 28.829 & 40.721 & 40.633 & 39.415 & 41.464 & 31.420 & 34.685 & 30.720  \\
 \cmidrule(lr){2-13}
 & CS $\uparrow$ & \textbf{0.9002} & \underline{0.8973} & 0.8606 & 0.8693 & 0.8541 & 0.8199 & 0.8207 & 0.8398 & 0.8601 & 0.8017 & 0.8596  \\
 & SS $\uparrow$ & \textbf{0.2613} & \underline{0.2599} & 0.2496 & 0.2447 & 0.2291 & 0.2289 & 0.2375 & 0.2217 & 0.2326 & 0.2379 & 0.2401  \\
\bottomrule
\end{tabular}%
}
\caption{Quantitative comparison with conventional and diffusion model baselines.}
\label{table:quantitative_comparison}
\end{table*}

\begin{figure*}[t]
    
    \centering
    \includegraphics[width=0.95\linewidth]{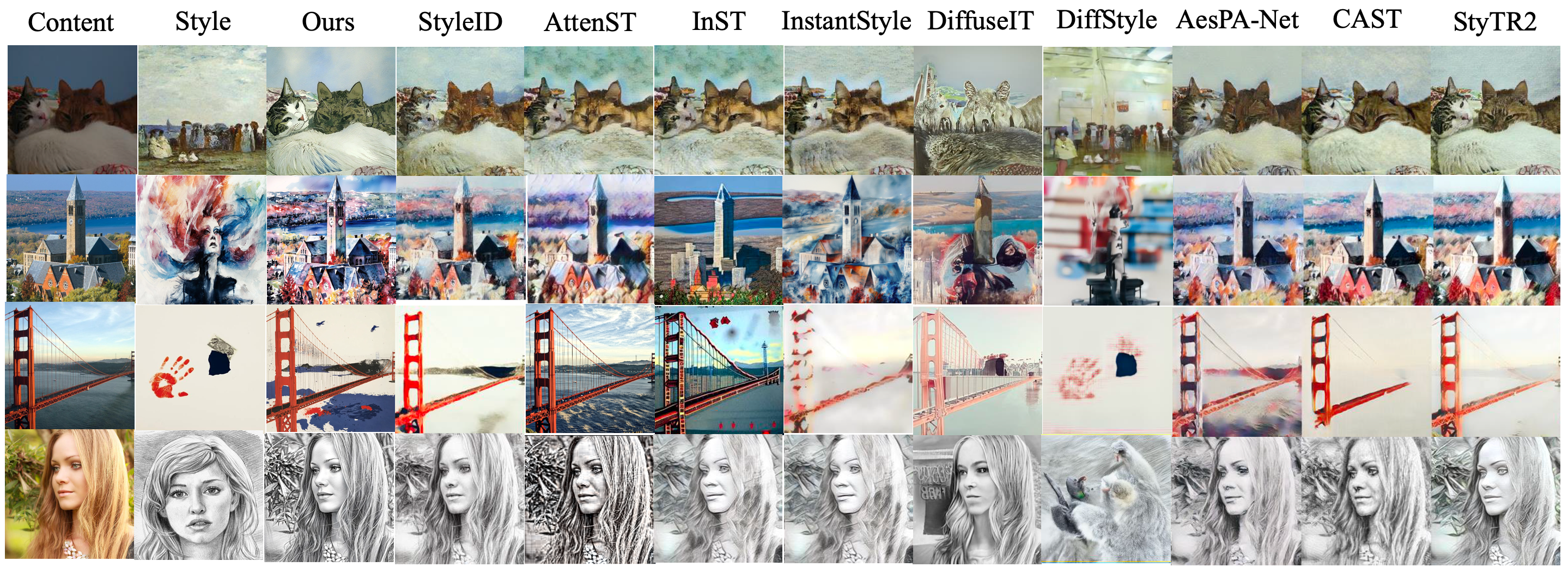}
    \caption{Qualitative comparison with conventional and diffusion model baselines.}
    \label{fig: Qualitative_compare}
\end{figure*}

%% file: sec/4_exp.tex
\begin{figure*}
    \centering
    \includegraphics[width=0.7\linewidth]{img/more_result_reduced.png}
    \caption{More results of photo-to-photo style transfer. The features presented in the style reference image—such as falling snowflakes, neon lights, blurred ground reflections, or sunset sunlight—are smoothly transferred with high aesthetic quality to the output image. }
    \label{fig:more_results}
\end{figure*}

\section{Experiments}
\subsection{Experimental setup}
\label{subsec: experimental setup}
We conduct all experiments using SD 1.5 and SDXL \cite{44} pretrained on the LAION dataset \cite{29}. Our model is fine-tuned on the PD12M dataset with 100,000 steps. To enable classifier-free guidance, we randomly drop text, content, and style images with a drop rate of $0.05$. The learning rate is set to $1e-4$. During the inference phase, we set denoising steps $t=50$ and the classifier-free guidance factor $CFG=10$.

\subsection{Evaluation Protocol}
\label{subsec:eval_protocal}
For artistic transfer tasks, traditional style transfer methods often rely on Style Loss \cite{35} as both a training objective and an evaluation metric, which can lead to results that overfit this loss. To ensure a more balanced comparison, we use the recently proposed ArtFID \cite{19} metric. ArtFID assesses overall style transfer performance by taking into account both content and style preservation and is known to correlate strongly with human judgment. ArtFID is calculated as $\text{ArtFID} = ( 1 + \text{LPIPS}_{content} )( 1 + \text{FID}_{style} )$. Here, $\text{LPIPS}_{content}$ \cite{26} evaluates the content fidelity between the stylized image and the original content image, while $\text{FID}_{style}$  \cite{18} measures the style fidelity between the stylized images and the style reference images. We conducted a quantitative comparison using 20 randomly selected content images from the MSCOCO \cite{17} dataset and 40 style images from WikiArt \cite{20}, generating 800 stylized images as StyleID has done. For photo-realistic transfer task, we selected 50 random content images from MSCOCO and 10 style images featuring natural styles such as sunrise, sunset, fog, snow, rain, autumn, and shadow. We used $\text{FID}_{style}$ , $\text{LPIPS}_{content}$, and Aesthetic Score for this quantitative comparison.

\subsection{Quantitative Comparison}
\label{subsec:quantitative_compare} 
Our method is evaluated through comparative analysis with nine leading style transfer approaches. These include 3 traditional methods: AesPA-Net \cite{21}, CAST \cite{22}, StyTR$^2$ \cite{23} and 6 diffusion-based techniques StyleID \cite{6}, AttenST \cite{52}, InST \cite{7}, InstantStyle \cite{16}, DiffuseIT \cite{46}, DiffStyle \cite{32}. For a fair comparison, we implemented all baselines using their publicly available code and recommended parameter settings.

As shown in Tab.~\ref{table:quantitative_comparison}, our model demonstrates exceptional performance in style transfer tasks, particularly excelling in photo-realistic transfers by outperforming conventional methods based on the Aesthetic Score. For artistic transfer tasks, we achieved the lowest ArtFID score, indicating superior style transfer with better content and style preservation that aligns closely with human judgment. Our method also leads in $\text{LPIPS}_{content}$ scores across all tasks, confirming strong content preservation along with effective style transfer. Additionally, Tab.~\ref{table:quantitative_comparison} reveals that the proposed method is highly compatible with various backbones, with a strong backbone significantly enhancing the model's results in both artistic and photo-realistic transfer tasks. 

\subsection{Qualitative Comparison}
\label{subsec: quanlitative comparision}
Fig.~\ref{fig: Qualitative_compare} demonstrates our method's superior content preservation while effectively transferring styles. Compared to StyleID (current SoTA), our model shows significantly better structural preservation—for instance, retaining bridge structure in the third row where StyleID and other baselines fail.

Photo-realistic transfer particularly highlights our advantages. Fig.~\ref{fig:Qualitative_compare_photo} reveals StyleID's limitations in capturing photo-realistic styles through self-attention layers, as styles like lighting and weather lack the distinct characteristics of artistic paintings. This causes inappropriate painting-like transformations (rows 1-2). While InST captures overall color changes, it fails with detailed elements like snow and shadows. Our approach excels at modifying both composition and specific details—adjusting shadows, lighting, and adding snowflakes—while preserving original content structure. Fig.~\ref{fig:more_results} shows additional photo-realistic transfer results.
\begin{figure*}[t]
    \centering
    \includegraphics[width=0.95\linewidth]{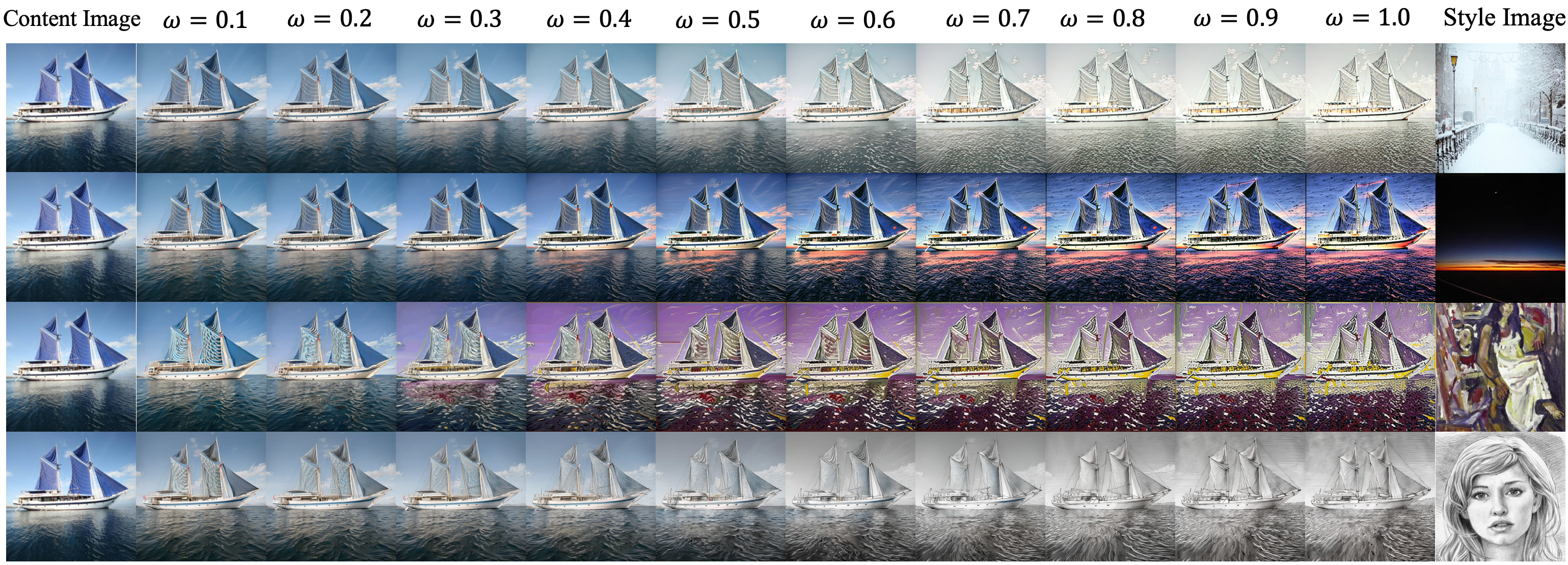}
    \caption{Continuously and smoothly control the influence level of a new style on the input image using CSAdaIN.}
    \label{fig:controllable_style_transfer}
\end{figure*}
\begin{figure*}
    \centering
    \includegraphics[width=0.95\linewidth]{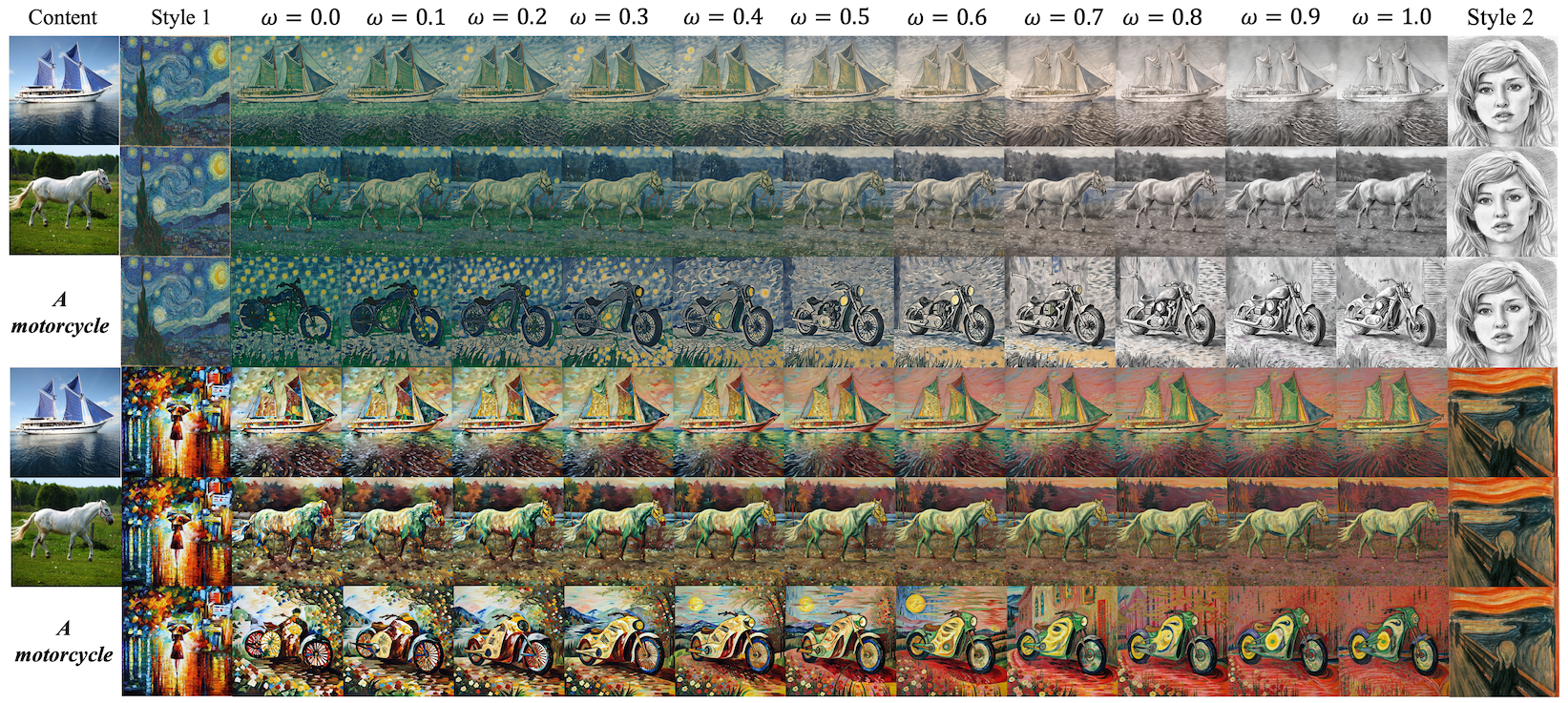}
    \caption{Style mixing with CSAdaIN for both style transfer and text-driven stylized synthesis tasks.}
    \label{fig:style_mixing}
\end{figure*}

\subsection{Evaluation of Content-Style Disentanglement}
Since our style extractor is created by subtracting content features (by content extractor) from CLIP features, evaluating one of the two components is sufficient. We choose to evaluate the style extractor and follow the exact same evaluation protocol as the Style Tokenizer \cite{49} for consistency and fairness.

We use 900 images across 12 style categories and compare our method with several existing style encoders: BlendGAN \cite{51}, Style Tokenizer \cite{49}, and InstantStyle \cite{16}. For each method, we extract style embeddings and perform clustering evaluation using two standard metrics: Silhouette Coefficient and Calinski-Harabasz.

These metrics measure how well features from the same style group together and how distinct they are from other groups. Results in  Tab.~\ref{table:style_extractor_eval} show that our Style Extractor achieves more compact and consistent clusters, outperforming other methods—especially the Style Tokenizer.
Furthermore, these results also support the choice of using 15 style categories in Section~\ref{subsec:content extractor}, indicating that this number is sufficient and appropriate for capturing stylistic diversity.

\begin{table}[ht]
    \centering
    \caption{Comparison with other style extraction methods}
    \begin{tabular}{lcc}
    \toprule
    \textbf{Method} & \textbf{Silhouette} $\uparrow$ & \textbf{Calinski-Harabasz} $\uparrow$ \\
    \midrule
    BlendGAN & 0.039 & 62.8 \\
    InstantStyle & 0.149 & 117.8 \\
    Style Tokenizer & 0.168 & 120.0 \\
    Ours (10 styles) & 0.153 & 114.2 \\
    Ours (15 styles) & \textbf{0.171} & \textbf{120.9} \\
    \bottomrule
    \end{tabular}
    \label{table:style_extractor_eval}
    \vspace{-10pt}
\end{table}

Additionally, content extractor \textit{\textbf{C}} and style extractor \textit{\textbf{S}} are designed to extract pure content and style features from any image. Therefore, they can be directly used to compute content similarity (\textbf{CS}) and style similarity (\textbf{SS}), as defined in Eq. 12 and 13.
\begin{align}
    CS &= cos(\boldsymbol{C}(I_c), \boldsymbol{C}(I_{cs})) \\
    SS &= cos(\boldsymbol{S}(I_s), \boldsymbol{S}(I_{cs}))
\end{align}

Results in Tab.~\ref{table:quantitative_comparison} show that \textbf{CS} correlates strongly with $\text{LPIPS}_{content}$, while \textbf{SS} aligns closely with $\text{FID}_{style}$. This indirect evidence suggests that both the content extractor and style extractor effectively capture their intended representations—pure content and style features.

\begin{table}[ht]
    \centering
    \caption{Inference time of diffusion-based methods}
    \resizebox{0.65\columnwidth}{!}{%
        \begin{tabular}{lcc}
        \toprule
        \textbf{Method} & \textbf{Output Size} & \textbf{Time (sec)} \\
        \midrule
        DiffuseIT & 512$\times$512 & 389.7 \\
        InST & 512$\times$512 & 404.8 \\
        DiffStyle & 512$\times$512 & 155.6 \\
        AttenST & 1024$\times$1024 & 8.31 \\
        StyleID & 512$\times$512 & 6.03 \\
        Ours (SD1.5) & 512$\times$512 & \textbf{1.89} \\
        Ours (SDXL) & 1024$\times$1024 & \textbf{4.45} \\
        \bottomrule
        \end{tabular}
    }
    \label{table:inference_time}
\end{table}

\subsection{Ablation Studies}
\label{subsec:abalation}
\noindent\textbf{Inference Time Comparison.} We measure inference time for content-style image pairs on a single A100 GPU (Tab.~\ref{table:inference_time}). Current methods require DDIM inversions or inference-stage optimization, adding significant overhead. StyleID, the fastest existing method, takes 6.03 seconds (4.02 for inversions, 2.01 for sampling). Our approach separates style-content features directly, enabling transfer from random noise without inversions. This achieves superior speed while maintaining fidelity—even SDXL at $1024\times1024$ resolution takes only 4.45 seconds.

\noindent\textbf{Ablation study on KVS Injection.} We replaced KVS Injection with conventional cross-attention for style feature injection, keeping other components unchanged. Tab.~\ref{tab:ablation_study_components} shows KVS Injection significantly improves $\text{FID}_{style}$ scores, confirming the importance of \textit{key} and \textit{value} features for style transfer. The slight decrease in $\text{LPIPS}_{content}$ reflects expected style-content trade-offs.

\noindent\textbf{Performance without content-style-stylized image triplet dataset.} 
Rows 5 and 10, Tab.~\ref{tab:ablation_study_components} show that without the content–style–stylized triplet dataset, our model loses a small degree of performance but still achieves state-of-the-art results. This highlights both the contribution of the style transfer consistency objective and the model’s effectiveness on standard datasets.

\noindent\textbf{Controllable style transfer.} Using Eq. 2 with varying $\omega$ values, Fig.~\ref{fig:controllable_style_transfer} demonstrates smooth style control across photo-realistic (normal-to-snowy, day-to-night) and artistic tasks (photo-to-paint, photo-to-sketch). Our framework enables multi-style synthesis (Fig.~\ref{fig:style_mixing}) by blending multiple target styles, while the content extractor eliminates unwanted original style influences.

\begin{table}[ht]
    \centering
    \caption{Ablation study of different configurations}
    \resizebox{\columnwidth}{!}{%
    \begin{tabular}{lccc}
    \toprule
    \multicolumn{4}{c}{\textbf{Artistic Style Transfer}} \\
    \midrule
    \textbf{Configuration} & \textbf{ArtFID} $\downarrow$ & \textbf{FID$_{\text{style}}$} $\downarrow$ & \textbf{LPIPS$_{\text{content}}$} $\downarrow$ \\
    \midrule
    Full method & \textbf{28.707} & \textbf{18.201} & 0.4951 \\
    w/o KVS Injection & 28.887 & 18.325 & \textbf{0.4948} \\
    w/o Style consistency loss & 28.914 & 18.338 & 0.4952 \\
    \midrule
    \multicolumn{4}{c}{\textbf{Photo-realistic Style Transfer}} \\
    \midrule
    \textbf{Configuration} & \textbf{Aesthetic Score} $\uparrow$ & \textbf{FID$_{\text{style}}$} $\downarrow$ & \textbf{LPIPS$_{\text{content}}$} $\downarrow$ \\
    \midrule
    Full method & \textbf{5.2198} & \textbf{22.417} & \textbf{0.3055} \\
    w/o KVS Injection & 5.2106 & 22.988 & \textbf{0.3055} \\
    w/o Style consistency loss & 5.1979 & 23.017 & 0.3085 \\
    \bottomrule
    \end{tabular}
    }
    \label{tab:ablation_study_components}
\end{table}
\vspace{-20pt}

%% file: sec/5_conclu.tex
\section{Conclusion}
In this paper, we introduce SCAdapter, a novel diffusion-based method for style transfer. Our approach effectively separates content features from the content image and style features from the style reference image, preventing cross-contamination and enabling cleaner, more authentic style transfers. We enhance this framework with three key innovations: CSAdaIN for precise multi-style blending with adjustable weighting, KVS Injection for targeted style integration, and a style transfer consistency objective to enhance overall style transfer and content preservation. Our experimental results demonstrate that SCAdapter outperforms existing methods, especially in photo-realistic transfer scenarios. Additionally, by eliminating the need for DDIM inversion and inference-stage optimization, our approach offers at least $2\times$ inference speed of previous diffusion-based methods.

%% file: main.bib
@String(CVPR= {IEEE Conf. Comput. Vis. Pattern Recog.})

@String(ICCV= {Int. Conf. Comput. Vis.})

@String(ECCV= {Eur. Conf. Comput. Vis.})

@String(CVPR  = {CVPR})

@String(ICCV  = {ICCV})

@String(ECCV  = {ECCV})


%% file: my.bib
@Article{1,
  author        = "Alex Nichol and Prafulla Dhariwal and Aditya Ramesh and Pranav Shyam and Pamela Mishkin and Bob McGrew and Ilya Sutskever and Mark Chen",
  title         = "Glide: Towards photorealistic image generation and editing with text-guided diffusion models",
  journal       = "arXiv preprint arXiv:2112.10741",
  year          = "2021",
  pages         = ""
}

@Article{2,
  author        = "Robin Rombach and Andreas Blattmann and Dominik Lorenz and Patrick Esser and Björn Ommer",
  title         = "High-resolution image synthesis with latent diffusion models",
  journal       = "Proceedings of the IEEE/CVF conference on computer vision and pattern recognition",
  year          = "2022",
  pages         = "10684--10695"
}

@Article{3,
  author        = "Omri Avrahami and Dani Lischinski and Ohad Fried",
  title         = "Blended diffusion for text-driven editing of natural images",
  journal       = "Proceedings of the IEEE/CVF Conference on Computer Vision and Pattern Recognition",
  year          = "2022",
  pages         = "18208--18218"
}

@Article{4,
  author        = "Wenhao Chai and Xun Guo and Gaoang Wang and Yan Lu",
  title         = "Stablevideo: Text-driven consistency-aware diffusion video editing",
  journal       = "Proceedings of the IEEE/CVF International Conference on Computer Vision",
  year          = "2023",
  pages         = "23040--23050"
}

@Article{5,
  author        = "Amir Hertz and Ron Mokady and Jay Tenenbaum and Kfir Aberman and Yael Pritch and Daniel Cohen-Or",
  title         = "Prompt-to-prompt image editing with cross-attention control",
  journal       = "The Eleventh International Conference on Learning Representations",
  year          = "2022",
  pages         = ""
}

@Article{6,
  author        = "Jiwoo Chung and Sangeek Hyun and Jae-Pil Heo",
  title         = "Style Injection in Diffusion: A Training-free Approach for Adapting Large-scale Diffusion Models for Style Transfer",
  journal       = "Proceedings of the IEEE/CVF Conference on Computer Vision and Pattern Recognition",
  year          = "2024",
  pages         = "8795--8805"
}

@Article{7,
  author        = "Yuxin Zhang and Nisha Huang and Fan Tang and Haibin Huang and Chongyang Ma and Weiming Dong and Changsheng Xu",
  title         = "Inversion-based style transfer with diffusion models",
  journal       = "Proceedings of the IEEE/CVF Conference on Computer Vision and Pattern Recognition",
  year          = "2023",
  pages         = "10146--10156"
}

@Article{8,
  author        = "Theofanis Karaletsos and Serge Belongie and Gunnar Rätsch",
  title         = "Bayesian representation learning with oracle constraints",
  journal       = "arXiv preprint arXiv:1506.05011",
  year          = "2015",
  pages         = ""
}

@Article{9,
  author        = "Haibo Chen and Lei Zhao and Huiming Zhang and Zhizhong Wang and Zhiwen Zuo and Ailin Li and Wei Xing and Dongming Lu",
  title         = "Diverse image style transfer via invertible cross-space mapping",
  journal       = "2021 IEEE/CVF International Conference on Computer Vision (ICCV)",
  year          = "2021",
  pages         = "14860--14869"
}

@Article{10,
  author        = "Gihyun Kwon and Jong Chul Ye",
  title         = "Diagonal attention and style-based gan for content-style disentanglement in image generation and translation",
  journal       = "Proceedings of the IEEE/CVF International Conference on Computer Vision",
  year          = "2021",
  pages         = "13980--13989"
}

@Article{11,
  author        = "Ian J. Goodfellow and Jean Pouget-Abadie and Mehdi Mirza and Bing Xu and David Warde-Farley and Sherjil Ozair and Aaron Courville and Yoshua Bengio",
  title         = "Generative adversarial nets",
  journal       = "Advances in Neural Information Processing Systems",
  year          = "2014",
  pages         = ""
}

@Article{12,
  author        = "Zhizhong Wang and Lei Zhao and Wei Xing",
  title         = "Stylediffusion: Controllable disentangled style transfer via diffusion models",
  journal       = "Proceedings of the IEEE/CVF International Conference on Computer Vision",
  year          = "2023",
  pages         = "7677--7689"
}

@Article{13,
  author        = "Hu Ye and Jun Zhang and Sibo Liu and Xiao Han and Wei Yang",
  title         = "IP-Adapter: Text Compatible Image Prompt Adapter for Text-to-Image Diffusion Models",
  journal       = "arXiv preprint arXiv:2308.06721",
  year          = "2023",
  pages         = ""
}

@Article{14,
  author        = "Alec Radford and Jong Wook Kim and Chris Hallacy and Aditya Ramesh and Gabriel Goh and Sandhini Agarwal and Girish Sastry and Amanda Askell and Pamela Mishkin and Jack Clark and Gretchen Krueger and Ilya Sutskever",
  title         = "Learning transferable visual models from natural language supervision",
  journal       = "International conference on machine learning",
  year          = "2021",
  pages         = "8748--8763"
}

@Article{15,
  author        = "Jordan Meyer and Nick Padgett and Cullen Miller and Laura Exline",
  title         = "Public Domain 12M: A Highly Aesthetic Image-Text Dataset with Novel Governance Mechanisms",
  journal       = "arXiv preprint arXiv:2410.23144",
  year          = "2024",
  pages         = ""
}

@Article{16,
  author        = "Haofan Wang and Matteo Spinelli and Qixun Wang and Xu Bai and Zekui Qin and Anthony Chen",
  title         = "InstantStyle: Free Lunch towards Style-Preserving in Text-to-Image Generation",
  journal       = "arXiv preprint arXiv:2404.02733",
  year          = "2024",
  pages         = ""
}

@Article{17,
  author        = "Tsung-Yi Lin and Michael Maire and Serge Belongie and Lubomir Bourdev and Ross Girshick and James Hays and Pietro Perona and Deva Ramanan and C. Lawrence Zitnick and Piotr Dollár",
  title         = "Microsoft coco: Common objects in context",
  journal       = "ECCV 2014",
  year          = "2014",
  pages         = "740--755"
}

@Article{18,
  author        = "Martin Heusel and Hubert Ramsauer and Thomas Unterthiner and Bernhard Nessler and Sepp Hochreiter",
  title         = "Gans trained by a two time-scale update rule converge to a local nash equilibrium",
  journal       = "Advances in neural information processing systems",
  year          = "2017",
  pages         = "30"
}

@Article{19,
  author        = "Matthias Wright, Björn Ommer",
  title         = "Artfid: Quantitative evaluation of neural style transfer",
  journal       = "DAGM German Conference on Pattern Recognition",
  year          = "2022",
  pages         = "560--576"
}

@Article{20,
  author        = "Wei Ren Tan and Chee Seng Chan and Hernan Aguirre and Kiyoshi Tanaka",
  title         = "Improved artgan for conditional synthesis of natural image and artwork",
  journal       = "IEEE Transactions on Image Processing",
  year          = "2019",
  pages         = "394--409"
}

@Article{21,
  author        = "Kibeom Hong and Seogkyu Jeon and Junsoo Lee and Namhyuk Ahn and Kunhee Kim and Pilhyeon Lee and Daesik Kim and Youngjung Uh and Hyeran Byun",
  title         = "Aespa-net: Aesthetic pattern-aware style transfer networks",
  journal       = "Proceedings of the IEEE/CVF International Conference on Computer Vision",
  year          = "2023",
  pages         = "22758--22767"
}

@Article{22,
  author        = "Yuxin Zhang and Fan Tang and Weiming Dong and Haibin Huang and Chongyang Ma and Tong-Yee Lee and Changsheng Xu",
  title         = "Domain enhanced arbitrary image style transfer via contrastive learning",
  journal       = "ACM SIGGRAPH 2022 Conference Proceedings",
  year          = "2022",
  pages         = "1--8"
}

@Article{23,
  author        = "Yingying Deng and Fan Tang and Weiming Dong and Chongyang Ma and Xingjia Pan and Lei Wang and Changsheng Xu",
  title         = "Stytr2: Image style transfer with transformers",
  journal       = "CVPR",
  year          = "2022",
  pages         = "11326--11336"
}

@Article{24,
  author        = "Yabin Zhang and Minghan Li and Ruihuang Li and Kui Jia and Lei Zhang",
  title         = "Exact feature distribution matching for arbitrary style transfer and domain generalization",
  journal       = "Proceedings of the IEEE/CVF Conference on Computer Vision and Pattern Recognition",
  year          = "2022",
  pages         = "8035--8045"
}

@Article{25,
  author        = "Xun Huang, Serge Belongie",
  title         = "Arbitrary style transfer in real-time with adaptive instance normalization",
  journal       = "Proceedings of the IEEE international conference on computer vision",
  year          = "2017",
  pages         = "1501--1510"
}

@Article{26,
  author        = "Richard Zhang and Phillip Isola and Alexei A. Efros and Eli Shechtman and Oliver Wang",
  title         = "The unreasonable effectiveness of deep features as a perceptual metric",
  journal       = "Proceedings of the IEEE conference on computer vision and pattern recognition",
  year          = "2018",
  pages         = "586--595"
}

@Article{27,
  author        = "Zhouxia Wang and Xintao Wang and Liangbin Xie and Zhongang Qi and Ying Shan and Wenping Wang and Ping Luo",
  title         = "Styleadapter: A single-pass lora-free model for stylized image generation",
  journal       = "arXiv preprint arXiv:2309.01770",
  year          = "2023",
  pages         = ""
}

@Article{28,
  author        = "Tianhao Qi and Shancheng Fang and Yanze Wu and Hongtao Xie and Jiawei Liu and Lang Chen and Qian He and Yongdong Zhang",
  title         = "Deadiff: An efficient stylization diffusion model with disentangled representations",
  journal       = "Proceedings of the IEEE/CVF Conference on Computer Vision and Pattern Recognition",
  year          = "2024",
  pages         = "8693--8702"
}

@Article{29,
  author        = "Christoph Schuhmann and Romain Beaumont and Richard Vencu and Cade Gordon and Ross Wightman and Mehdi Cherti and Theo Coombes and Aarush Katta and Clayton Mullis and Mitchell Wortsman and others",
  title         = "Laion-5b: An open large-scale dataset for training next generation image-text models",
  journal       = "Advances in Neural Information Processing Systems",
  year          = "2022",
  pages         = "25278--25294"
}

@Article{30,
  author        = "Chitwan Saharia and William Chan and Saurabh Saxena and Lala Li and Jay Whang and Emily L. Denton and Kamyar Ghasemipour and Raphael Gontijo Lopes and Burcu Karagol Ayan and Tim Salimans and others",
  title         = "Photorealistic text-to-image diffusion models with deep language understanding",
  journal       = "Advances in Neural Information Processing Systems",
  year          = "2022",
  pages         = "36479--36494"
}

@Article{31,
  author        = "Martin Nicolas Everaert and Marco Bocchio and Sami Arpa and Sabine Süsstrunk and Radhakrishna Achanta",
  title         = "Diffusion in style",
  journal       = "Proceedings of the IEEE/CVF International Conference on Computer Vision",
  year          = "2023",
  pages         = "2251--2261"
}

@Article{32,
  author        = "Jaeseok Jeong and Mingi Kwon and Youngjung Uh",
  title         = "Training-free style transfer emerges from h-space in diffusion models",
  journal       = "arXiv preprint arXiv:2303.15403",
  year          = "2023",
  pages         = ""
}

@Article{33,
  author        = "Zhizhong Wang and Lei Zhao and Wei Xing",
  title         = "StyleDiffusion: Controllable Disentangled Style Transfer via Diffusion Models",
  journal       = "Proceedings of the IEEE/CVF International Conference on Computer Vision",
  year          = "2023",
  pages         = "7677--7689"
}

@Article{34,
  author        = "Serin Yang and Hyunmin Hwang and Jong Chul Ye",
  title         = "Zero-shot Contrastive Loss for Text-Guided Diffusion Image Style Transfer",
  journal       = "arXiv preprint arXiv:2303.08622",
  year          = "2023",
  pages         = ""
}

@Article{35,
  author        = "Leon A Gatys and Alexander S Ecker and Matthias Bethge",
  title         = "Image Style Transfer Using Convolutional Neural Networks",
  journal       = "Proceedings of the IEEE Conference on Computer Vision and Pattern Recognition",
  year          = "2016",
  pages         = "2414--2423"
}

@Article{36,
  author        = "Wei-Sheng Lai and Jia-Bin Huang and Narendra Ahuja and Ming-Hsuan Yang",
  title         = "Deep Laplacian Pyramid Networks for Fast and Accurate Super-Resolution",
  journal       = "Proceedings of the IEEE Conference on Computer Vision and Pattern Recognition",
  year          = "2017",
  pages         = "624--632"
}

@Article{37,
  author        = "Yijun Li and Chen Fang and Jimei Yang and Zhaowen Wang and Xin Lu and Ming-Hsuan Yang",
  title         = "Universal Style Transfer via Feature Transforms",
  journal       = "Advances in Neural Information Processing Systems",
  year          = "2017",
  pages         = ""
}

@Article{38,
  author        = "Huan Wang and Yijun Li and Yuehai Wang and Haoji Hu and Ming-Hsuan Yang",
  title         = "Collaborative Distillation for Ultra-Resolution Universal Style Transfer",
  journal       = "Proceedings of the IEEE/CVF Conference on Computer Vision and Pattern Recognition",
  year          = "2020",
  pages         = "1860--1869"
}

@Article{39,
  author        = "Zhizhong Wang and Lei Zhao and Haibo Chen and Lihong Qiu and Qihang Mo and Sihuan Lin and Wei Xing and Dongming Lu",
  title         = "Diversified Arbitrary Style Transfer via Deep Feature Perturbation",
  journal       = "Proceedings of the IEEE/CVF Conference on Computer Vision and Pattern Recognition",
  year          = "2020",
  pages         = "7789--7798"
}

@Article{41,
  author        = "Xiaohua Zhai and Joan Puigcerver and Alexander Kolesnikov and Pierre Ruyssen and Carlos Riquelme and Mario Lucic and Josip Djolonga and Andre Susano Pinto and Maxim Neumann and Alexey Dosovitskiy and Lucas Beyer and Olivier Bachem and Michael Tschannen and Marcin Michalski and Olivier Bousquet and Sylvain Gelly and Neil Houlsby",
  title         = "A Large-scale Study of Representation Learning with the Visual Task Adaptation Benchmark",
  journal       = "arXiv preprint arXiv:1910.04867",
  year          = "2020",
  pages         = ""
}

@Article{42,
  author        = "Lvmin Zhang and Anyi Rao and Maneesh Agrawala",
  title         = "Adding Conditional Control to Text-to-Image Diffusion Models",
  journal       = "Proceedings of the IEEE/CVF International Conference on Computer Vision",
  year          = "",
  pages         = "3836--3847"
}

@Article{43,
  author        = "Berel Lang",
  title         = "The Concept of Style",
  journal       = "Cornell University Press",
  year          = "1987",
  pages         = "4"
}

@Article{44,
  author        = "Dustin Podell and Zion English and Kyle Lacey and Andreas Blattmann and Tim Dockhorn and Jonas Müller and Joe Penna and Robin Rombach",
  title         = "SDXL: Improving Latent Diffusion Models for High-Resolution Image Synthesis",
  journal       = "arXiv preprint arXiv:2307.01952",
  year          = "2023",
  pages         = ""
}

@Article{45,
    author = "Amir Hertz and Ron Mokady and Jay Tenenbaum and Kfir Aberman and Yael Pritch and Daniel Cohen-Or",
    title = "Prompt-to-Prompt Image Editing with Cross Attention Control",
    journal       = "arXiv preprint arXiv:2208.01626",
    year          = "2023",
    pages         = ""

}

@Article{46,
  author        = "Gihyun Kwon and Jong Chul Ye",
  title         = "Diffusion-based Image Translation Using Disentangled Style and Content Representation",
  journal       = "arXiv preprint arXiv:2209.15264",
  year          = "2022",
  pages         = ""
}

@Article{47,
  author        = "Serin Yang and Hyunmin Hwang and Jong Chul",
  title         = "Zero-shot contrastive loss for text-guided diffusion image style transfer",
  journal       = "arXiv preprint arXiv:2303.08622",
  year          = "2023",
  pages         = ""
}

@article{49,
    author = "Wen Li and Muyuan Fang, Cheng Zou and Biao Gong and Ruobing Zheng and Meng Wang and Jingdong Chen and Ming Yang",
    title = "StyleTokenizer: Defining Image Style by a Single Instance for Controlling Diffusion Models",
    journal = "European Computer Vision Association (ECCV)",
    year = "2024",
}

@misc{51,
      title={BlendGAN: Implicitly GAN Blending for Arbitrary Stylized Face Generation}, 
      author={Mingcong Liu and Qiang Li and Zekui Qin and Guoxin Zhang and Pengfei Wan, Wen Zheng},
      year={2021},
      eprint={2110.11728},
      archivePrefix={arXiv},
      primaryClass={cs.CV}
}

@misc{52,
      title={AttenST: A Training-Free Attention-Driven Style Transfer Framework with Pre-Trained Diffusion Models}, 
      author={Bo Huang and Wenlun Xu and Qizhuo Han and Haodong Jing and Ying Li},
      year={2025},
      eprint={2503.07307},
      archivePrefix={arXiv},
      primaryClass={cs.CV}
}
